\documentclass[a4paper,conference]{IEEEtran}
\usepackage{cite}
\usepackage{amsmath,amssymb,amsfonts}
\usepackage{algorithmic}
\usepackage{graphicx}
\usepackage{caption}
\usepackage{subfig}
\usepackage{textcomp}
\usepackage{epsfig}
\usepackage{dsfont}
\usepackage{calrsfs}
\usepackage{threeparttable}
\usepackage[linesnumbered,ruled,vlined]{algorithm2e}
\usepackage{mathtools}
\usepackage{array}
\usepackage{booktabs}
\usepackage{bm} 
\usepackage{multirow}
\usepackage{enumerate}
\usepackage{enumitem}
\usepackage{makecell}
\usepackage[subnum]{cases}
\usepackage{dsfont}
\usepackage{fancyhdr}
\graphicspath{ {IMAGES/} }
\usepackage{balance}
\usepackage[pagebackref=false,breaklinks=true,letterpaper=true,colorlinks,bookmarks=false]{hyperref}

\makeatletter
\newcommand*\titleheader[1]{\gdef\@titleheader{#1}}
\AtBeginDocument{%
  \let\st@red@title\@title
  \def\@title{%
    \bgroup\normalfont\normalsize\centering\vspace{-1.2cm}\@titleheader\par\egroup
    \vskip1em\st@red@title}
}
\makeatother

\title{One-shot Representational Learning for Joint Biometric and Device Authentication}
\titleheader{\textcolor{red}{S. Banerjee and A. Ross, ``One-shot Representational Learning for Joint Biometric and Device Authentication," Accepted in 25th International Conference on Pattern Recognition (ICPR), (Milan, Italy), January 2021}}
\author{\IEEEauthorblockN{Sudipta Banerjee and Arun Ross}
\IEEEauthorblockA{Michigan State University \\
Email: \{banerj24, rossarun\}@cse.msu.edu}
}

\begin{document}

%



\maketitle

\begin{abstract}
In this work, we propose a method to simultaneously perform (i) biometric recognition (\textit{i.e.}, identify the individual), and (ii) device recognition, (\textit{i.e.}, identify the device) from a single biometric image, say, a face image, using a one-shot schema. Such a joint recognition scheme can be useful in devices such as smartphones for enhancing security as well as privacy. We propose to automatically learn a joint representation that encapsulates both biometric-specific and sensor-specific features. We evaluate the proposed approach using iris, face and periocular images acquired using near-infrared iris sensors and smartphone cameras. Experiments conducted using 14,451 images from 15 sensors resulted in a rank-1 identification accuracy of upto 99.81\% and a verification accuracy of upto 100\% at a false match rate of 1\%.  
\end{abstract}
\IEEEpeerreviewmaketitle

\section{Introduction}

Biometric data such as face, fingerprint or iris images reveal information about the identity of the individual as well as the identity of the device used to acquire the data~\cite{BiomSensor_1, BiomSensor_2}. In some applications such as smartphone banking, it is necessary to authenticate both the {\em user} as well as the {\em device} in order to enhance security~\cite{Galdi_PRL_15, Ross_ICB_18}. This can be done by invoking two separate modules: one for biometric recognition and the other for device or sensor recognition.\footnote{The terms ``device" and ``sensor" are interchangeably used in this paper. Thus, determining the identity of a smartphone camera (\textit{i.e.,} sensor) is akin to determining the identity of the smartphone (\textit{i.e.,} device) itself.} In such cases, the system has to store two distinct templates: a biometric template denoting the identity of the user and sensor template denoting the identity of the device.  

In this paper, we approach this problem by designing a {\em joint template} that can be used to authenticate both the user and the device simultaneously. Our objective is as follows: \textit{Given a biometric image we would like to simultaneously recognize the individual and the acquisition device}. In the process of accomplishing this objective, we address the following questions:

\begin{figure}[t]
\centering
\subfloat[]
{   
    \includegraphics[scale=.37]{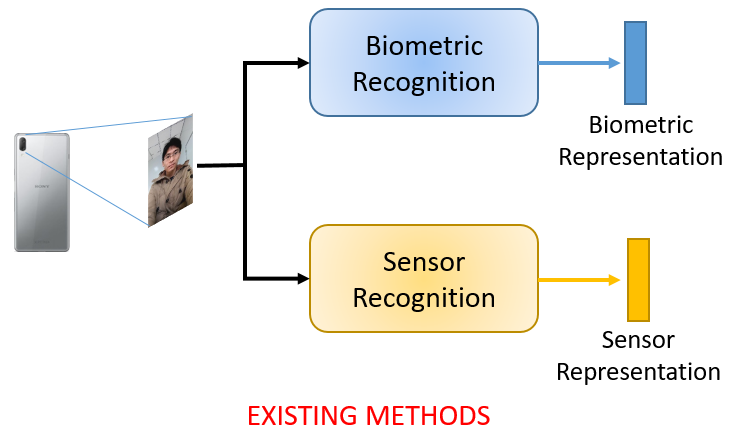} 
}\\
\subfloat[]
{ 
    \includegraphics[scale=.37]{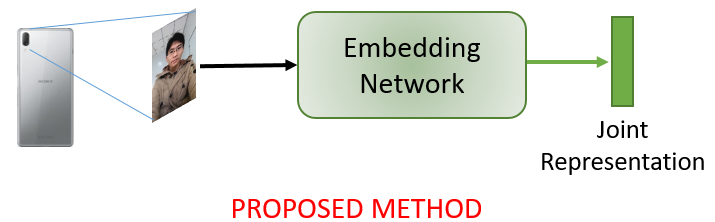} 
}

\caption{Difference between (a) methods that use separate modules for computing biometric and sensor representations, and (b) the proposed method that uses an embedding network to generate a joint biometric-sensor representation. }
\label{Fig: Obj}
\end{figure}
\begin{enumerate}
\item \textbf{Why do we need to combine biometric and device recognition?} \\ 
Smartphones are increasingly using biometrics for access control and monetary transactions. Examples include fingerprint and face recognition on iPhones and iris recognition on Samsung Galaxy S9. Device verification\footnote{Typically used in two factor authentication (2FA)  protocol that combines any two of the three factors: `something you are' (biometrics), `something you have' (a code on the authorized device) and `something you know' (a password) for additional security.} can provide assurance that the biometric sample is being acquired by an authorized device. A combined biometric and device recognition system can therefore guarantee that the right person is accessing the remote service (\textit{e.g.}, banking) using an authorized device. 

\item \textbf{Can existing device verification techniques be used in the smartphone application scenario?} \\
Device identification can be performed using the MAC (media access control) address, a unique networking address assigned to each device. However, in case of smartphones that have multiple network interfaces, such as Wi-Fi, 4G, bluetooth, etc., there can be multiple MAC addresses which may be broadcasted, making them vulnerable. Alternatively, SRAM cells can be used to deduce physically unclonable cues for device identification~\cite{Ross_ICB_18}; this is a hardware-based solution and requires access to the physical device. In a mobile banking scenario, where the verification is conducted remotely, the customer provides a biometric sample in the form of an image, and some device information, but not necessarily the physical device itself. In this scenario, hardware-based solutions will be ineffective. 

\item \textbf{Why do we need a joint representation?} \\
Existing literature uses separate modules to tease out the biometric-specific and sensor-specific details from an image and perform feature-level or score-level fusion~\cite{Galdi_PRL_15, Ross_ICB_18}. However, they suffer from the following limitations: (i) the overall performance is limited by the weakest recognition module, and (ii) the process may not generalize well across different biometric modalities and multi-spectral sensors. Therefore, a \textit{joint} representation that combines both biometric and sensor-specific features present in a biometric image can offer the following advantages: (i) the joint representation is not constrained by the performance of the individual recognition module, and the same \textit{method} can be employed across different biometric modalities, and (ii) the joint representation integrates the biometric and sensor representations into a compact template, such that, the  individual templates cannot be easily de-coupled; this implicitly imparts privacy to the biometric component.

\end{enumerate} 

%
%
%

\section{Related Work}

Biometric recognition systems comprise a \textit{feature extraction} module that elicits a salient feature representation from the acquired biometric data, and a \textit{comparator} module that compares two sets of feature representations to compute a match score~\cite{Jain_Intro}. On the other hand, sensor recognition systems extract sensor pattern noise~\cite{Lukas_TIFS_06} from a set of training images obtained from different sensors to generate \textit{sensor reference patterns}. To deduce the sensor identity of an unknown test image, first its sensor pattern noise is extracted, and then it is \textit{correlated} with the reference patterns. The test image is assigned to the device whose reference pattern yields the highest correlation value.

In~\cite{Galdi_ICIAP}, the authors used partial face images acquired using smartphones and employed a weighted sum fusion rule at the score level to combine sensor and biometric recognition. Later, they extended their work to include feature level fusion in~\cite{Galdi_PRL_15} and concluded that score level fusion performed comparatively better. In~\cite{Galdi_Face}, the authors performed HOG-based face recognition and combined it with Photo Response Non-Uniformity-based sensor recognition at the score level. In~\cite{Ross_ICB_18}, the authors combined fingerprint recognition with device recognition by performing feature level fusion of minutiae-cylinder-codes with SRAM start-up values. Fusion at the score or feature level is often dependent on the specific biometric modality and the device sensor used. A specific fusion rule producing the best results on a particular biometric and sensor modality (\textit{e.g,} iris and near-infrared sensors) may not yield optimal results on a different modality (\textit{e.g,} face and RGB sensors), and therefore, needs to be tuned separately for each pair of biometric and sensor modalities. Furthermore, feature-level fusion retains the individual biometric and sensor-specific components that can be recovered from the fused representation using appropriate measures. Obtaining the biometric template may compromise the privacy aspect of biometrics. In contrast, the proposed joint representation non-trivially unifies the biometric and sensor-specific features. As a result, typical countermeasures will be ineffective in disentangling the biometric component from the joint representation. This will implicitly preserve the privacy of the biometric component.


The remainder of the paper is organized as follows. Section~\ref{Sec:Prop} describes the proposed method. Section~\ref{Sec:DSandExp} describes the dataset and experimental protocols used in this work. Section~\ref{Sec:Res} reports the results. Section~\ref{Sec:Sum} summarizes the findings and concludes the paper.

\section{Motivation and Proposed Method}
\label{Sec:Prop}
An image contains both low frequency and high frequency components. For example, in a face image, the low frequency components capture the illumination details while the high frequency components capture the structural details present in the face that are useful for biometric recognition. Recently, sensor recognition has been successully accomplished using Photo Response Non-Uniformity (PRNU)~\cite{Lukas_TIFS_06} for different types of sensors, such as DSLR sensors~\cite{Chen_TIFS_08}, smartphone sensors~\cite{BTAS_19}, and also near-infrared iris sensors~\cite{Uhl4_ICB_12}. PRNU is a form of sensor pattern noise in an image that manifests due to anomalies during the fabrication process and is, therefore, unique to each sensor. Typically, PRNU resides in the high frequencies that can be useful for sensor recognition~\cite{Lukas_TIFS_06}. Since the high frequencies dominate in both biometric and sensor representations, we hypothesize that there is a joint representation, that, if effectively extracted, can be utilized for both tasks of biometric and sensor recognition. Our objective  is to \textit{learn} this joint representation that lies at the intersection of the sensor and biometric space. Mathematically, it can be represented as 
$\displaystyle J(\bm{X}) = B(\bm{X}) \cap S(\bm{X})$, where $\bm{X}$ is an input biometric image, $B(\cdot)$ is the biometric representation extracted from $\bm{X}$, $S(\cdot)$ is the sensor representation computed from the same input $\bm{X}$, and $J(\cdot)$ is the joint representation. Existing methods process $\bm{X}$ using two independent routines to extract the two representations, and can optionally perform fusion, either at feature level or at score level, to make a decision. However, we propose to leverage the two representations to derive a joint representation (see Figure~\ref{Fig: Obj}). The joint space can be best approximated using an embedding network that can convert images to compact representations~\cite{Emb1}. The embedding network  $\mathcal{E}$, takes two inputs, $\bm{X}$ and the dimensionality ($k$) of the embedding to be generated, such that $\displaystyle  J(\bm{X}) = \mathcal{E}(\bm{X},k) \approx B(\bm{X}) \cap S(\bm{X})$. The second argument $k$, allows us to regulate the dimensionality of the joint representation, which will be much lesser than the original dimensionality of the image, as well as the combined dimensionality of the two representations computed separately, \textit{i.e.}, if $\bm{X} \in \mathbb{R}^d$, $B(\bm{X}) \in \mathbb{R}^m$ and $S(\bm{X}) \in \mathbb{R}^n$, then the joint representation $J(\bm{X}) \in \mathbb{R}^k$, where, $k << d$ and $k < (m+n)$.

\begin{figure*}[h]
\centering

    \includegraphics[scale=.39]{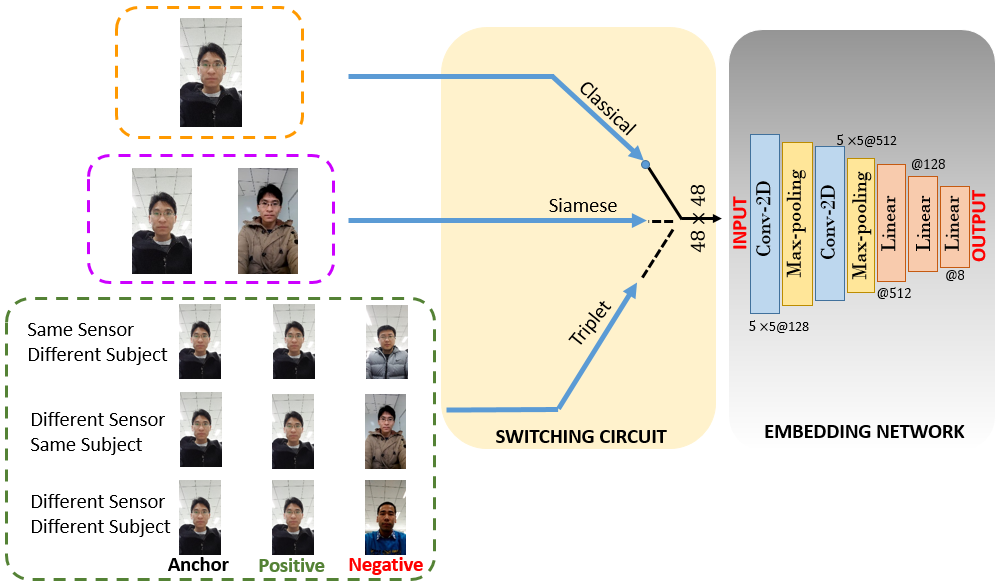} 

\caption{Outline of the proposed method used for computing the joint biometric and sensor representation. \textbf{Input:} A single image, or a pair of images, or 3-tuple images to the embedding network. \textbf{Output:} Joint biometric-sensor representation. The embedding network is trained in three mutually exclusive modes, \textit{viz.,} classical mode (top row), siamese mode (middle row) and triplet mode (bottom row). The switching circuit selects only one training mode at a time.}
\label{Fig: Prop}
\end{figure*}

In this work, we used a deep convolutional neural network that serves the role of the embedding network (see Figure~\ref{Fig: Prop}). The embedding network consists of two 2-D convolutional layers and three linear layers. We used max-pooling for down-sampling the feature map and a parametric-rectified linear activation unit (PReLU) as the activation function. The embedder accepts an image, resized to $48 \times 48$, as the input and produces a 8-dimensional output, which is the \textit{joint} representation. The choice of the dimensionality of the representation along with the experimental setup is described later (see Section~\ref{Sec: Expts}).

The \textbf{main contributions} of this work are as follows:
\begin{enumerate}
\item We propose a method to learn a joint biometric and sensor representation using a one-shot approach that can be used in joint identification and joint verification scenarios. A correct joint identification/verification occurs \textit{only} if both subject identity and device identity yield correct matches.
\item We employ an embedding network that can learn the joint representation irrespective of the biometric modality and the sensor used for acquisition. In this context, we evaluate the proposed method using three different biometric modalities (face, iris and periocular), and different types of sensors (iris sensors operating in the near-infrared spectrum and smartphone camera sensors operating in the visible spectrum).
\item We perform extensive experiments using different training paradigms and loss functions, and compare the proposed method with existing state-of-the-art algorithms for biometric and sensor recognition.
\end{enumerate}

\section{Datasets and Experiments}
\label{Sec:DSandExp}
\subsection{Datasets}

In this work, we focused on three different biometric modalities, \textit{viz}., iris, periocular and face. To this end, we used three different datasets - (i) CASIA-Iris Image Database Version 2~\cite{CasV2} which contains near-infrared iris images acquired using two sensors, (ii) Mobile Iris Challenge Evaluation (MICHE-I) dataset~\cite{MICHE} which contains partial face images acquired using two smartphones (front and rear sensors separately) and front camera of a tablet, and (iii) OULU-NPU dataset~\cite{OULU} which contains face images acquired using the front sensors of six smartphones. We used \textit{only} bonafide images from all three sessions in the OULU-NPU dataset. Table~\ref{Tab: Dataset} describes the datasets used in this work. Note that the smartphone datasets (MICHE-I and OULU-NPU) contain images acquired in the visible spectrum.

\begin{table*}[]
\caption{Dataset specifications used in this work. We used three datasets corresponding to 3 biometric modalities \textit{viz.,} iris, periocular and face. Here, we perform joint biometric and sensor recognition, so total $\# Classes$ is computed as the product of $\# Subjects$ and $\# Sensors$. ($^*$MICHE-I dataset has a total 75 subjects, out of which the first 48 subjects were imaged using iPhone 5S UNIT I and the remaining 27 subjects were imaged using iPhone 5S UNIT II, as observed in~\cite{Galdi_PRL_15}. Here, `UNIT' refers to two different units of the same brand and model iPhone 5S, and therefore, should be treated as two different smartphones. In this case, $\# Classes$ = 375 since only a subset of the total 75 subjects were imaged using either of the two units of iPhone 5S smartphone at a time. Therefore, $75 \times 3$ [Samsung] $+ 48 \times 2$ [UNIT I] $+ 27 \times 2$ [UNIT II] $= 375$.) }
\label{Tab: Dataset}
\begin{tabular}{|l|ll|l|l|l|} \hline
\textbf{Modality}           & \textbf{Dataset}               & \textbf{\begin{tabular}[c]{@{}l@{}}Name of sensors\end{tabular}}                                                                                                  & \textbf{\begin{tabular}[c]{@{}l@{}}(\# Subjects, \# Sensors, \\ \# Classes)\end{tabular}} & \textbf{Split} & \textbf{\# Images} \\ \hline \hline
\multirow{2}{*}{Iris}       & \multirow{2}{*}{CASIA-Iris V2} & \multirow{2}{*}{CASIA IrisCAM-V2, OKI IrisPass-h}                                                                                                                   & \multirow{2}{*}{(60, 2, 120)}                                                             & Train          & 1,680              \\
                            &                                &                                                                                                                                                                     &                                                                                           & Test           & 720                \\ \hline
\multirow{2}{*}{Periocular} & \multirow{2}{*}{MICHE-I}       & \multirow{2}{*}{\begin{tabular}[c]{@{}l@{}}Apple iPhone 5S (Front and Rear) UNIT I and UNIT II, \\ Samsung Galaxy S4 (Front and Rear), Samsung Galxy Tab GT2 (Front)\end{tabular}}     & \multirow{2}{*}{(75, 7, 375$^*$)}                                                             & Train          & 2,278              \\
                            &                                &                                                                                                                                                                     &                                                                                           & Test           & 863                \\ \hline
\multirow{2}{*}{Face}       & \multirow{2}{*}{OULU-NPU}      & \multirow{2}{*}{\begin{tabular}[c]{@{}l@{}}HTC Desire EYE, Sony XPERIA C5 Ultra Dual, MEIZU X5,\\ Oppo N3, Samsung Galaxy S6 Edge, ASUS Zenfne Selfie\end{tabular}} & \multirow{2}{*}{(55, 6, 330)}                                                             & Train          & 5,940              \\
                            &                                &                                                                                                                                                                     &                                                                                           & Test           & 2,970              \\ \hline \hline
\multicolumn{3}{|l|}{TOTAL}                                                                                                                                                                                                          & (190, 15, 825)                                                                            &                & 14,451            \\ \hline
\end{tabular}
\end{table*}

\subsection{Evaluation Protocol}
\label{Sec: EvalProto}
Before we describe the experiments, we present the protocol that is used to evaluate the proposed approach. We evaluate the method in two scenarios, \textit{viz.,} (i) joint identification and (ii) joint verification. The terms joint identification and joint verification are different from the terms used conventionally in the biometric literature. In the case of \textbf{joint identification}, a correct identification occurs \textit{only} when both sensor and subject labels of the test sample match with the ground truth labels. To perform evaluation in the joint identification scenario, we select one embedding from each class (combines both sensor and subject label) to form the gallery, and the remaining embeddings are used as probes. We use two metrics to compute the distance or similarity between the probe and gallery embeddings and select the top three matches: (i) standardized Euclidean distance (computes the pairwise euclidean distance divided by the standard deviation) and (ii) cosine similarity. We plot the cumulative match characteristics (CMC) curves corresponding to the top three ranks. In the case of \textbf{joint verification}, two joint representations will yield a match if both the embeddings belong to the same sensor and same subject, otherwise a mismatch occurs. Incorrect match can occur in three cases as follows: (i) if the two joint representations belong to the same subject, but different sensors, (ii) if the two joint representations belong to the same sensor, but different subjects, and (iii) if the two joint representations belong to different subjects and different sensors. To perform evaluation in the joint verification scenario, we compute the distance or similarity between all the test embeddings and present receiver operating characteristics (ROC) curves to indicate the joint verification performance. We also report the true match rate (TMR) values @1\% and 5\% false match rates (FMR). 

\subsection{Experimental Settings}
\label{Sec: Expts}

In this work, we designed the experimental settings using three different modes of training. See Figure~\ref{Fig: Prop}. Say, $\bm{O}$ denotes the output of an embedding network for input $\bm{X}$, \textit{i.e.}, $\bm{O} = \mathcal{E}(\bm{X},k)$. In the first mode, referred to as the \textit{classical} mode, the embedding $\bm{O}$ is fed to a classification network which minimizes the cross-entropy loss computed between the ground truth label and the predicted label. The classification network in our case is a shallow network which applies PReLU activation on the embedding, followed by a fully-connected layer and then applies softmax to compute a probability value. We assigned the ground truth label for the $i^{th}$ image, such that $l_i \in Sub_i \otimes\ Sen_i$, where $Sub_i$ denotes the subject identifier of image $i$, $Sen_i$ denotes the sensor identifier for the same image and $\otimes$ denotes the tensor product. The cardinality of the set of labels $|L| = |Sub \times Sen|$. In the second mode, referred to as the \textit{siamese} mode, a siamese network~\cite{siamese} is used which feeds a pair of images to the embedding network. The embedding network then computes a pair of embeddings ($\bm{O}_i, \bm{O}_j$) and the siamese network is trained by minimizing the contrastive loss~\cite{contrastive} computed between the pair of embeddings. We used single margin (SMCL) and double margin (DMCL) contrastive losses. Finally, in the third mode, referred to as the \textit{triplet} mode, a triplet network~\cite{tripletnetwork} is trained using embeddings generated from an anchor ($\bm{O}_a$), a positive ($\bm{O}_p$) and a negative ($\bm{O}_n$) sample by minimizing the triplet loss~\cite{tripletloss}. We performed offline triplet mining as well as online triplet mining~\cite{TLmining} with different triplet selection strategies (random negative triplet selection, semi hard negative triplet selection and hardest negative triplet selection). The triplet loss considers only one negative example at a time. Alternatively, multi-class N-pair loss function~\cite{npair} considers multiple negative instances from several classes. In this work, we consider a positive example as one which belongs to the same class as the anchor (same subject and same sensor), whereas there can be three types of negative examples, \textit{viz}., same subject but different sensor, same sensor but different subject and different sensor with different subject. Therefore, the number of negative classes in this work is significantly high, so we used multi-class N-pair loss using two mining techniques: (i) all positive pairs and (ii) hard negative pairs. Table~\ref{Tab: Loss} summarizes the different loss functions used in the three training modes in this work. Note that each input to the embedding network as shown in Figure~\ref{Fig: Prop} is \textbf{mutually exclusive}, \textit{i.e.}, the embedding network can operate independently in any of the three training modes. We modified the design of an existing embedding network for implementing the different training paradigms~\cite{imp}. We used learning rate = $1\times \exp{(-4)}$, batch size = 4, Adam optimizer, and a step decay to reduce the learning rate by a factor $\gamma = 0.1$ every 8 epochs. The proposed network is shallow so we trained only for 50 epochs. The margin values in single margin contrastive loss and triplet losses are set to 1, while in double margin contrastive loss, both margins are set to 0.5.


\begin{table}[]
\centering
\caption{Description of the training modes and the loss functions used in this work.}
\label{Tab: Loss}
\scalebox{1.1}{
\begin{tabular}{|l|l|l|} \hline
\multicolumn{1}{|c|}{\textbf{Training mode}} & \multicolumn{2}{c|}{\textbf{Loss function}}                  \\  \hline \hline
Classical                                  & \multicolumn{2}{c|}{Cross entropy}                           \\ \hline
\multirow{2}{*}{Siamese}                   & \multicolumn{2}{c|}{Single margin contrastive loss (SMCL)}          \\
                                           & \multicolumn{2}{c|}{Double margin contrastive loss (DMCL)}          \\ \hline
\multirow{6}{*}{Triplet}                   & \multicolumn{2}{c|}{Offline triplet mining}                  \\ \cline{2-3}
                                           & \multirow{3}{*}{Online triplet mining} & Random negative    \\
                                           &                                        & Semi-hard negative \\
                                           &                                        & Hardest negative   \\ \cline{2-3}
                                           & \multirow{2}{*}{Multi-class N-pair}    & All positive pair  \\
                                           &                                        & Hard negative pair \\ \hline
\end{tabular}}
\end{table}


For each dataset, we used a training set and a test set (see Table~\ref{Tab: Dataset}). The number of classes is computed as the product of the number of sensors and number of subjects in that dataset. For example, CASIA-Iris V2 dataset has 60 subjects and 2 sensors, so total number of classes is $60 \times 2=120$. Each class has 20 images, therefore, the total number of images (samples) is 2,400 ($20 \times 120$). The training and test partitions follows a 70:30 split. So, for a single class, out of 20 samples, 14 samples are randomly selected as the training set and the remaining 6 samples form the test set. Similar protocol is followed for the remaining datasets.         

\begin{figure}[h]
\centering

    \includegraphics[scale=.27]{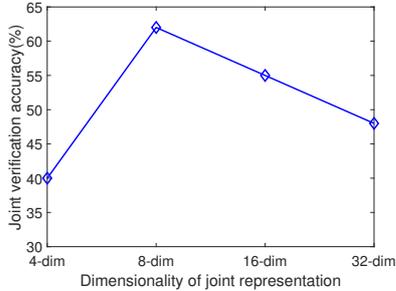} 

\caption{Variation in the joint verification performance as a function of the dimensionality of the joint representation. Experiment is conducted on the validation set using 50 images from the MICHE-I dataset and four dimensionality values \textit{viz.,} $\{4,8,16,32\}$. 8-dimensional embedding resulted in the highest joint verification accuracy, and is therefore selected in this work.}
\label{Fig: Dim}
\end{figure} 

Next, in the training phase, the embedding network accepts an image (resized to $48 \times 48$) as input. Different image resolutions were used $\{28 \times 28, 48 \times 48, 96 \times 96 \}$, but $48 \times 48$ provided optimal trade-off between accuracy and training time. The embeddings are trained in (i) classical, (ii) siamese and (iii) triplet modes. Then, in the testing phase, we computed the embeddings from the test set. We evaluate the test embeddings in joint identification and joint verification scenarios. 

Although deep learning-based sensor identification methods exist in the literature~\cite{CR_Ref1, CR_Ref2, CR_Ref3}, we used Enhanced PRNU~\cite{Li_TIFS_10} (with enhancement Model III) as the sensor identification baseline for all three modalities due to its low computational burden and effectiveness against multi-spectral images~\cite{Ross_18}. Enhanced PRNU requires creation of sensor reference patterns, that serve as gallery and test (probe) noise residuals, that are correlated with the reference patterns. We used training images to compute the sensor reference patterns and test images for correlation. A test image is assigned to the sensor class resulting in the highest correlation value. See~\cite{BTAS_19} for more details. Test noise residuals computed from JPEG images can be matched successfully against sensor reference patterns computed from RAW images~\cite{PRNU_JPEG}, thereby, justifying the use of PRNU as a state-of-the-art sensor identification baseline. We used COTS matcher as the biometric recognition baseline for iris and face modalities. For the periocular modality, we used a pretrained ResNet-101 architecture~\cite{ResNet} and used the features from layer 170 as the biometric representation for the test samples. This particular architecture is used because it has demonstrated good performance in biometric verification on the MICHE-I dataset~\cite{BTAS_19}. The gallery comprises the training images and the probes are the test images. Since, PRNU can only be used for the task of sensor identification, we selected to implement both the baselines only in identification scenario.

We further conducted an experiment using a validation set comprising 50 images from the MICHE-I dataset (excluded from the test set) to analyze the effect of the dimensionality of the embedding on the verification performance. To this end, we used four values of $k = \{4,8,16,32\}$, and then selected that value which results in the highest performance for the remaining experiments.

\begin{figure}[]
\centering

    \includegraphics[scale=.27]{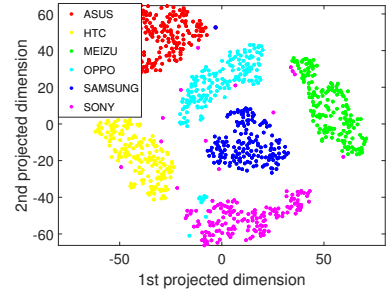} 

\caption{2-D projection of the embeddings using t-SNE used for sensor identification in the OULU-NPU dataset. Each sensor class is sufficiently discriminated from the rest of the sensors.}
\label{Fig: tsne}
\end{figure}

\section{Results and Analysis}
\label{Sec:Res}

\subsection{Selection of the metric and dimensionality of embedding}

In terms of the choice of the distance/similarity \textbf{metric}, we observed that standardized euclidean distance metric resulted in better performance compared to the cosine similarity metric. This can be attributed to the standardization process which takes into account the intra-class and inter-class variations in the embeddings. In terms of the choice of the \textbf{dimensionality} of the embedding, we observed that 8 was the optimal value, since, it resulted in the best performance (64\% on the MICHE-I validation set) as indicated in Figure~\ref{Fig: Dim}. Therefore, we used 8-dimensional embedding and standardized Euclidean distance metric for all the experiments. Furthermore, we presented the t-SNE~\cite{TSNE} visualization of the performance of the embedding network in terms of sensor identification for the OULU-NPU dataset in Figure~\ref{Fig: tsne}. The well-separable clusters corresponding to the six sensors demonstrate the capability of the embedding network used in this work.

\subsection{Performance of each of the three training modes}

\begin{figure*}[h]
\centering
\subfloat[]
{   
    \includegraphics[scale=.3]{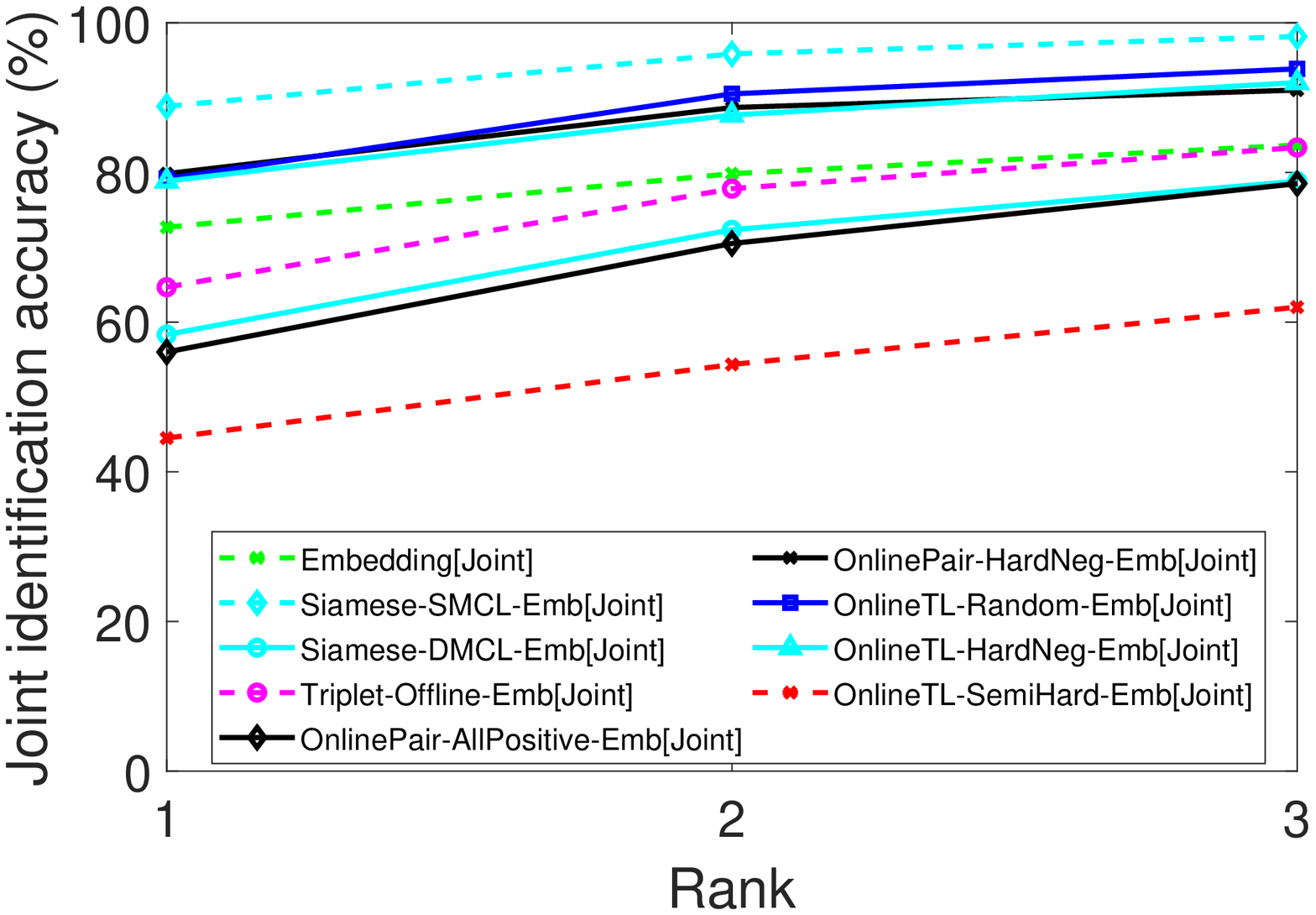} 
} \hspace{-0.75cm}
\subfloat[]
{ 
    \includegraphics[scale=.3]{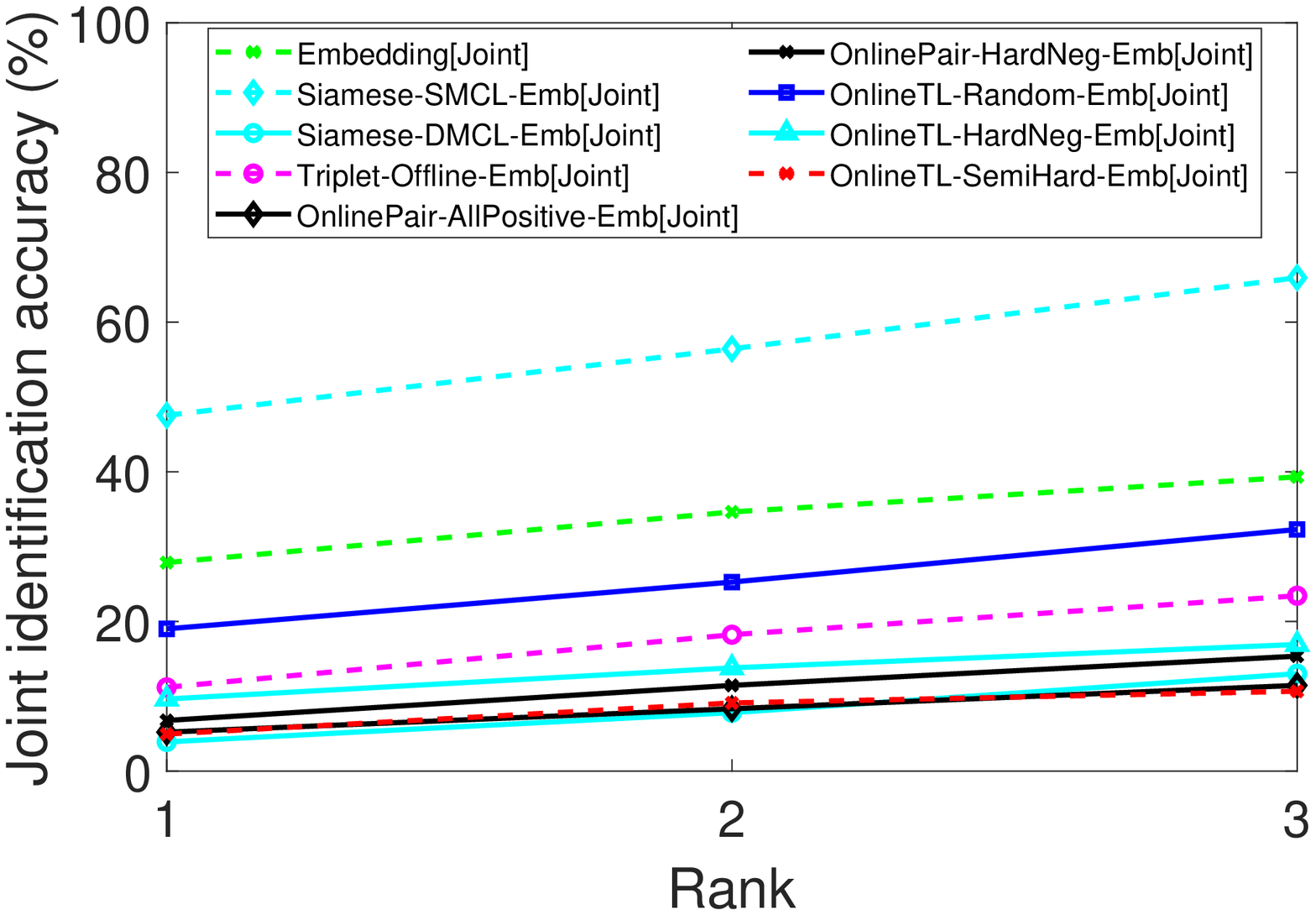} 
} \hspace{-0.75cm}
\subfloat[]
{ 
    \includegraphics[scale=.3]{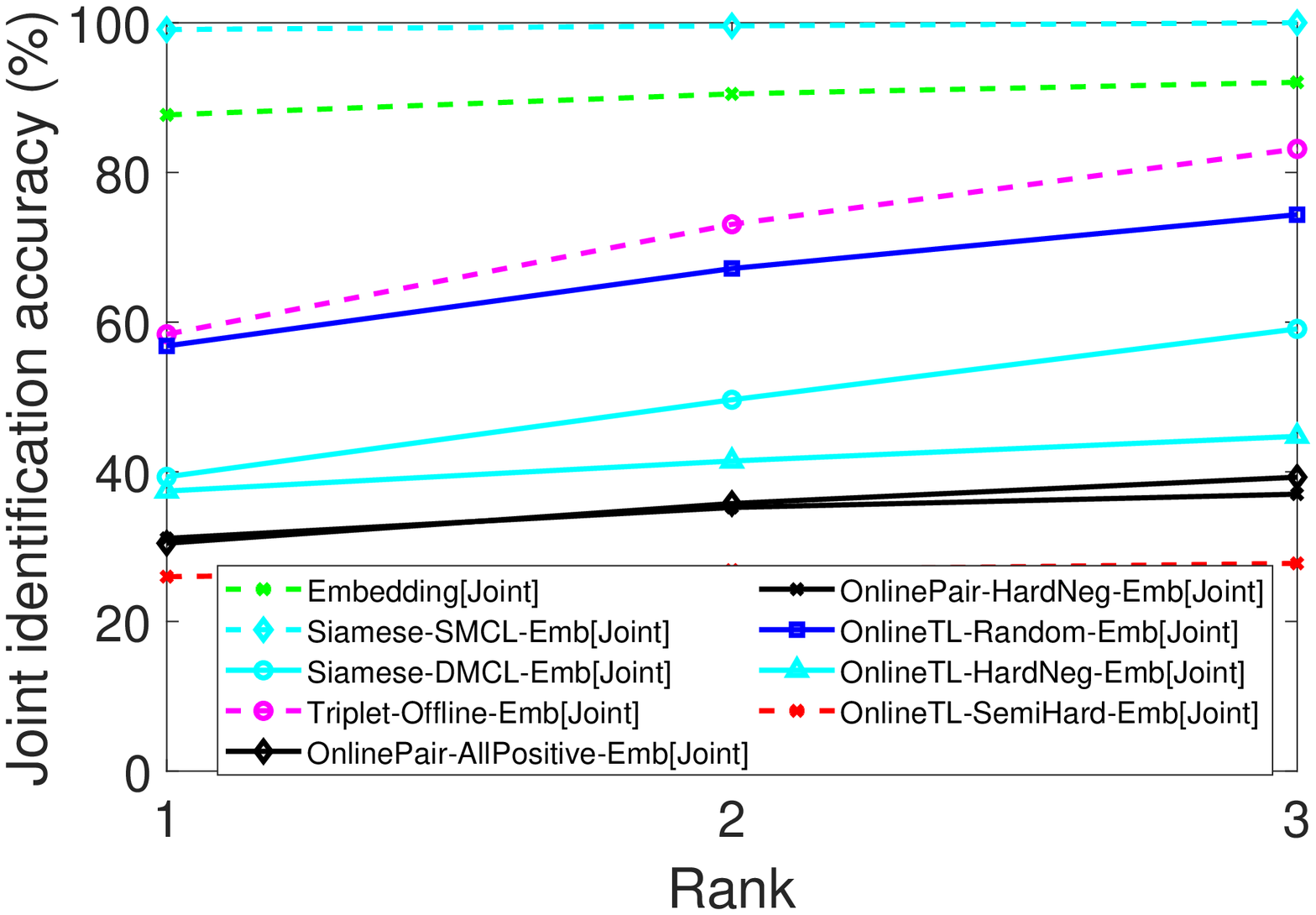}   
}

\caption{Cumulative Matching Characteristics (CMC) curves for the proposed method in the \textbf{joint identification scenario} for the following datasets: (a) CASIA-Iris V2 (b) MICHE-I and (c) OULU-NPU. Refer to Table~\ref{Tab: Loss} for the different training networks and loss functions indicated in the legend in an identical order.}
\label{Fig: CMC}
\end{figure*}

\begin{figure*}[h]
\centering
\subfloat[]
{   
    \includegraphics[scale=.315]{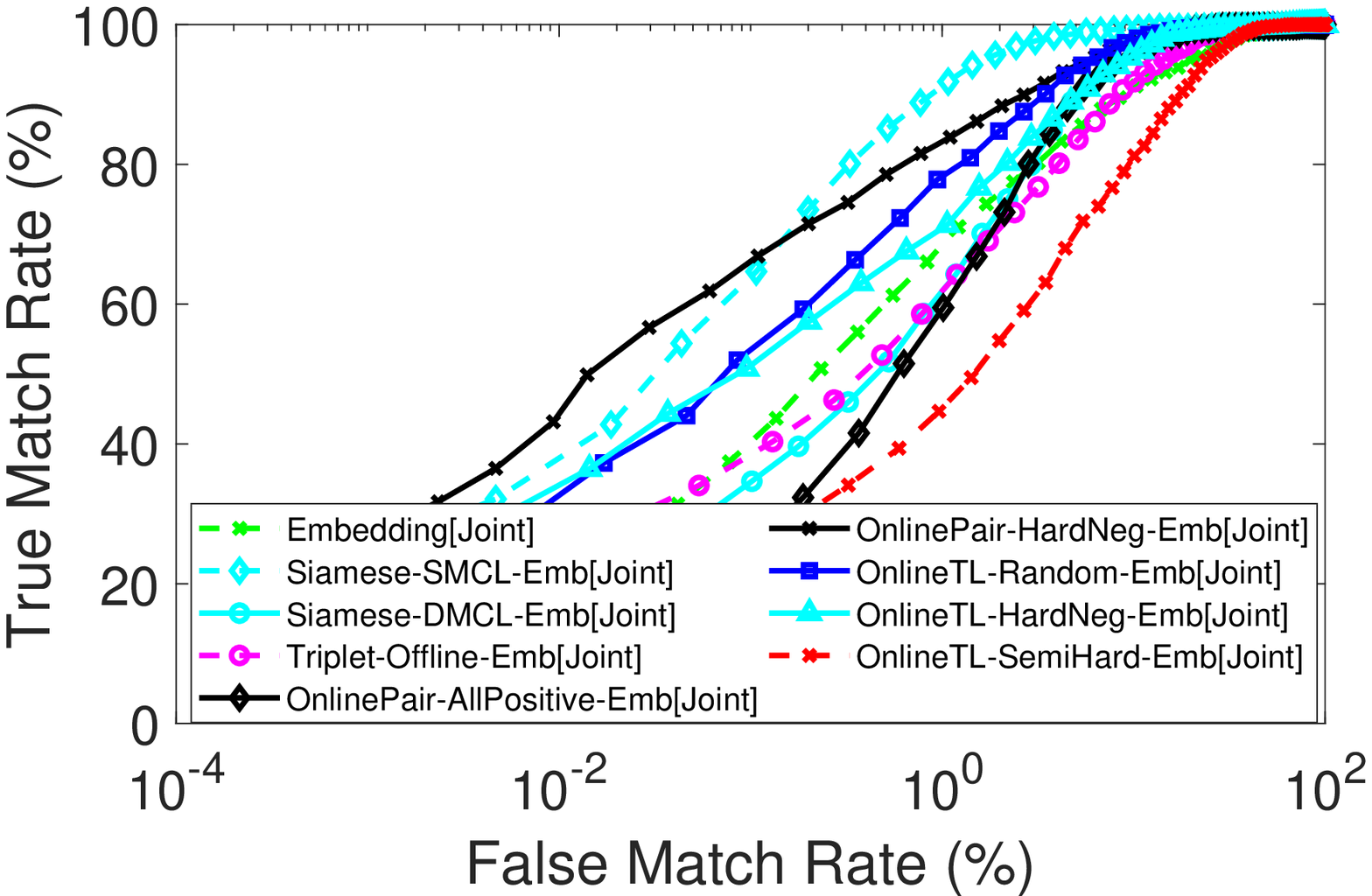} 
} \hspace{-0.75cm}
\subfloat[]
{ 
    \includegraphics[scale=.315]{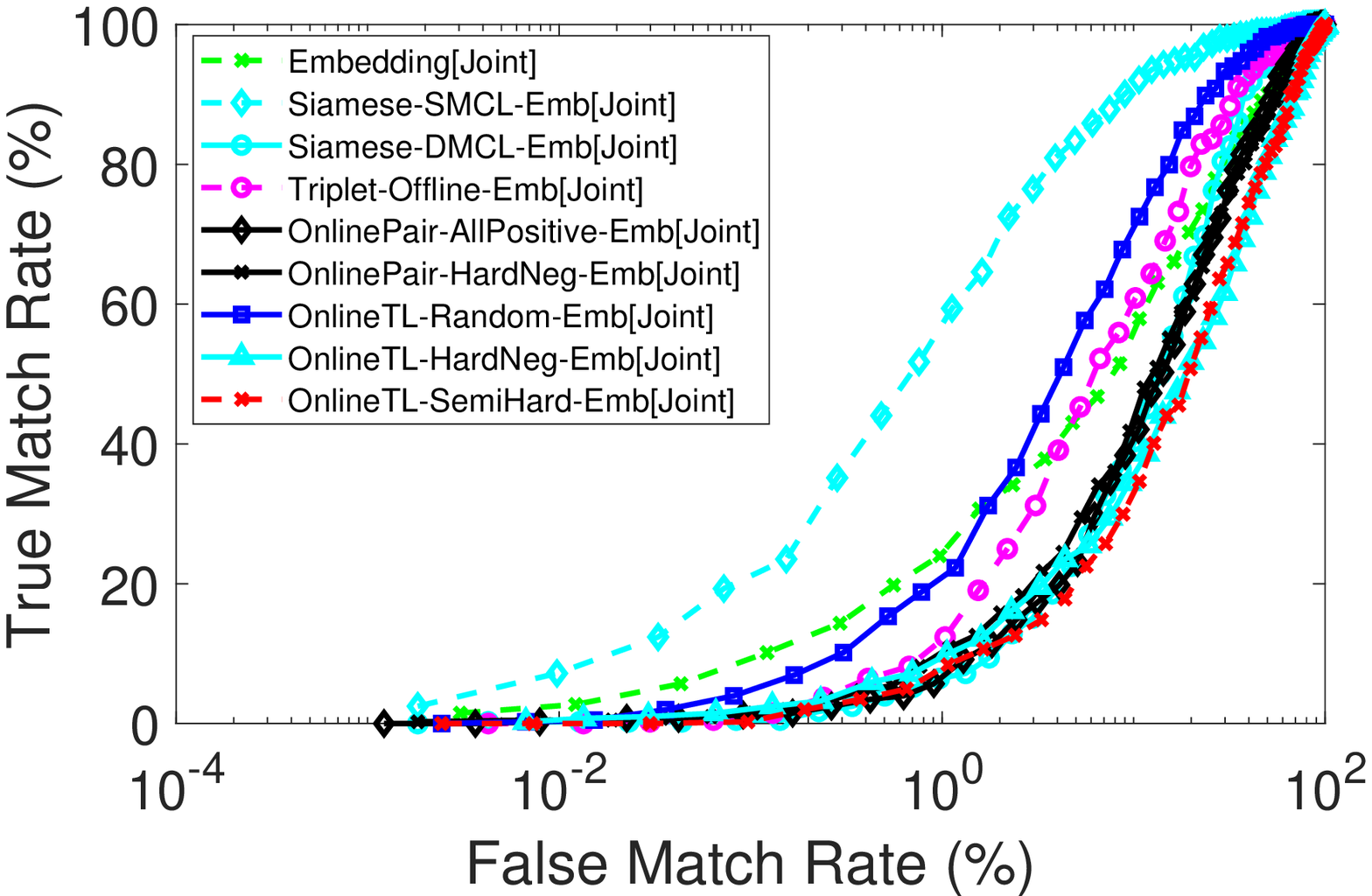} 
}\hspace{-0.75cm}
\subfloat[]
{ 
    \includegraphics[scale=.315]{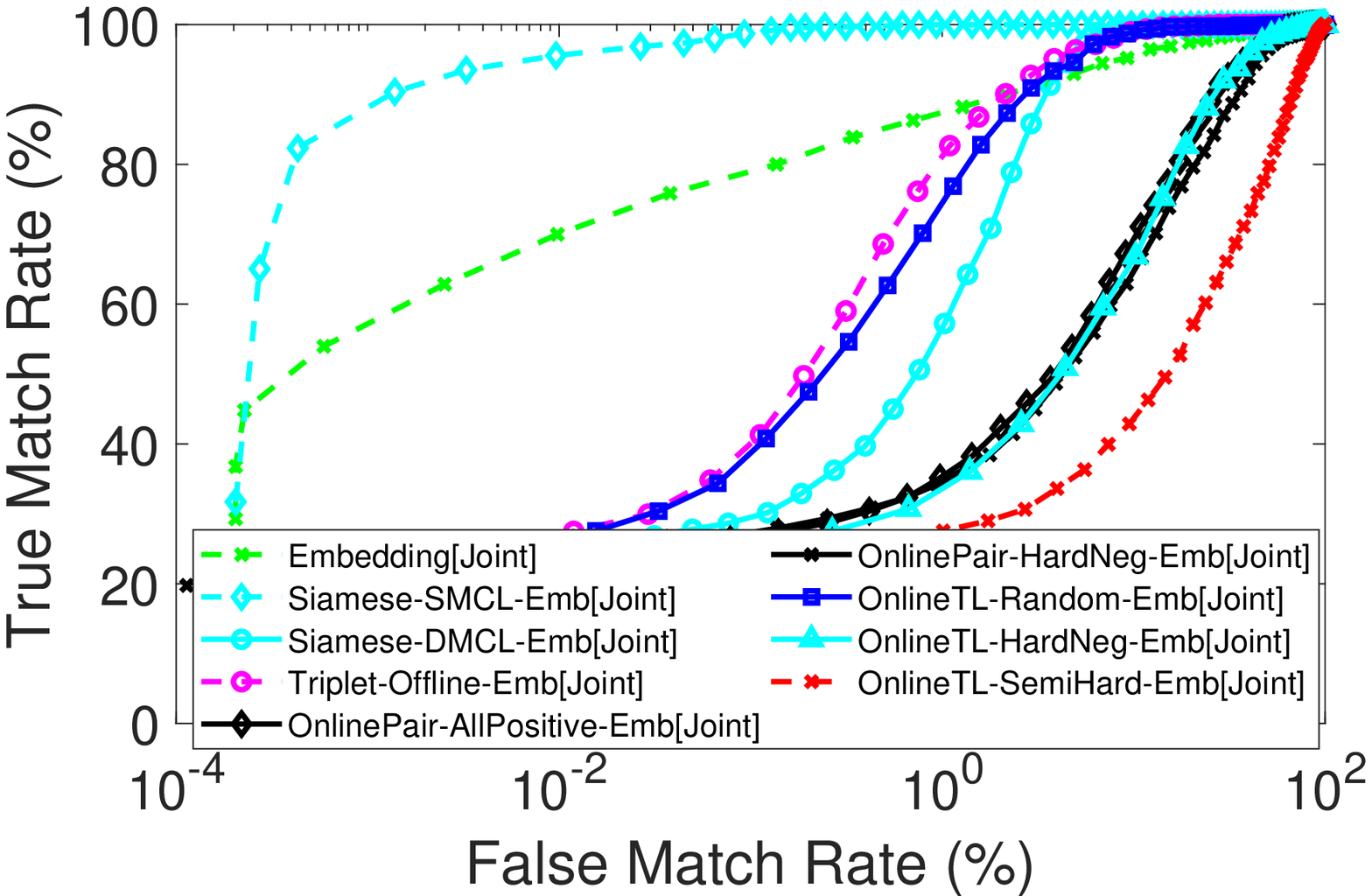}   
}

\caption{Receiver Operating Characteristics (ROC) curves for the proposed method in the \textbf{joint verification scenario} for the following datasets: (a) CASIA-Iris V2 (b) MICHE-I and (c) OULU-NPU. Refer to Table~\ref{Tab: Loss} for the different training networks and loss functions indicated in the legend in an identical order. }
\label{Fig: ROC}
\end{figure*}

In terms of \textbf{training algorithms}, the overall results in both joint identification and joint verification scenarios indicate that the embedding network trained in \textit{siamese} mode outperformed the remaining training paradigms (see Figures~\ref{Fig: CMC} and~\ref{Fig: ROC}). The reason for the superior performance of siamese network can be attributed to the use of contrastive loss. Out of the two contrastive losses, single margin contrastive loss outperformed double margin contrastive loss. The contrastive loss considers a pair of embeddings at a time, and tries to either minimize the distance between them if they belong to the same class, or increases the distance between them by some margin if they belong to different classes. On the other hand, triplet loss tries to \textit{simultaneously} minimize the distance between the anchor and positive sample, whereas, maximize the distance between the anchor and negative samples. In this work, the number of negative classes is very high (in a 330 class dataset, 1 class is positive and the remaining 329 classes are negative). This makes the task of triplet loss much more complex as compared to contrastive loss. Given the huge variation in the possible combination of negative triplets (see Figure~\ref{Fig: Prop}), we suspect that the triplet loss struggled to determine the accurate decision boundary between the positive and negative classes, resulting in an overall reduction in performance. We investigated different types of triplet mining strategies, and observed that online triplet mining outperformed offline triplet mining and multi-class N-pair in a majority of the cases. 

\begin{table}[]
\centering
\caption{Results in the joint identification scenario. Results are reported in terms of Rank 1 identification accuracies (\%). A correct joint identification implies that \textit{both} sensor and subject resulted in a match. Mismatch of either subject or sensor or both will result in an incorrect joint identification.}
\label{Tab: Ident}
\scalebox{0.8}{
\begin{tabular}{|l|l|l|l|l|l|} \hline
\multirow{2}{*}{Dataset} & \multirow{2}{*}{\begin{tabular}[c]{@{}l@{}}Method for \\ baseline\end{tabular}} & \multicolumn{3}{c|}{Baseline performance}                                                                                                                                                            & \multirow{2}{*}{\begin{tabular}[c]{@{}l@{}}Proposed \\ method \\ (\%)\end{tabular}} \\ \cline{3-5}
                         &                                                                                          & \begin{tabular}[c]{@{}l@{}}Sensor\\ identification \\ (\%) \end{tabular} & \begin{tabular}[c]{@{}l@{}}Biometric\\ identification \\ (\%)\end{tabular} & \begin{tabular}[c]{@{}l@{}}Joint\\ identification \\ (\%)\end{tabular} &                                                                                        \\ \hline \hline
\begin{tabular}[c]{@{}l@{}}CASIA- \\Iris V2 \end{tabular}            & \begin{tabular}[c]{@{}l@{}}PRNU \\COTS \end{tabular}  & 100.00                                                          & 56.52                                                            & 56.52                                                          & \textbf{89.67 } \\ \hline
MICHE-I                  & \begin{tabular}[c]{@{}l@{}}PRNU \\ResNet-101 \end{tabular}  & 99.86                                                           & 18.05                                                            & 18.05                                                          & \textbf{47.53                                                                                 } \\ \hline
OULU-NPU                 & \begin{tabular}[c]{@{}l@{}}PRNU \\COTS \end{tabular}  & 98.48                                                           & 84.24                                                            & 83.13                                                        & \textbf{99.81 } \\ \hline  
\end{tabular}}
\end{table}

\subsection{Results of the joint identification experiment}
In terms of the performance in \textbf{joint identification scenario}, Table~\ref{Tab: Ident} compares the results with the baseline performance for all the datasets. We reported the baselines for sensor identification (PRNU), biometric identification (COTS or ResNet), followed by joint identification, separately. We reiterate that joint identification involves a correct match only if both sensor and subject labels are correct to allow fair comparison with the proposed method. Results indicate that the proposed method outperformed the baseline (joint identification) by 26.41\% averaged across all three datasets computed at Rank 1. The poor performance for the MICHE-I dataset can be attributed to two factors - firstly, the large number of classes (= 375) compared to rest of the datasets (see Table~\ref{Tab: Dataset}), and secondly, the diverse acquisition settings (indoor vs. outdoor) resulting in degraded biometric recognition, and subsequently leading to overall poor performance. Surprisingly, the proposed method can still outperform the baseline by $\sim$30\%. We have further analyzed this performance in Section~\ref{MICHE_Analysis}. CMC curves indicate the superior performance of the siamese network in contrast to classical and triplet networks. 

\begin{table}[]
\centering
\caption{Results in the joint verification scenario. Results are reported in terms of true match rate (TMR) at false match rates (FMRs) of 1\% and 5\%.}
\label{Tab: Veri}
\scalebox{1.2}{
\begin{tabular}{|l|l|l|} \hline
Dataset       & TMR@FMR=1\% & TMR@FMR=5\% \\ \hline \hline
CASIA-Iris V2 & 90.00     & 98.00     \\ \hline
MICHE-I       & 62.00     & 90.00     \\ \hline 
OULU-NPU      & 100.00    & 100.00 \\ \hline  
\end{tabular}}
\end{table}

\subsection{Results of the joint verification experiment}
In terms of the performance in \textbf{joint verification scenario}, Table~\ref{Tab: Veri} reports the results. Results indicate that the proposed method achieved an average joint TMR of 84\% @1\% FMR, and an average TMR of 96\%  @5\% FMR, indicating the strong representative capability of the joint representation. ROC curves in Figure~\ref{Fig: ROC} indicate that the joint representation learnt using siamese network trained with single margin contrastive loss (see the curve marked Siamese-SMCL-Emb[Joint]) outperformed the remaining joint representations. We would like to point out that in~\cite{Galdi_PRL_15}, the authors achieved 23\% (by using feature level fusion) and 86\% (by using score level fusion) at 5\% FMR on the MICHE-I dataset (the authors excluded the Samsung Galaxy Tab 2 subset of the MICHE-I dataset, which we included in our evaluations). Although their objectives were different compared to the proposed work (they adopted a fusion rule for integrating their proposed biometric and sensor recognition performances), we would like to indicate that the task of joint recognition is difficult. In spite of that, the proposed method performed reasonably well.

\subsection{Analysis of the performance of the proposed method on MICHE-I dataset}
\label{MICHE_Analysis}
\begin{figure}[h]
\centering
\subfloat[]
{   
    \includegraphics[scale=.07]{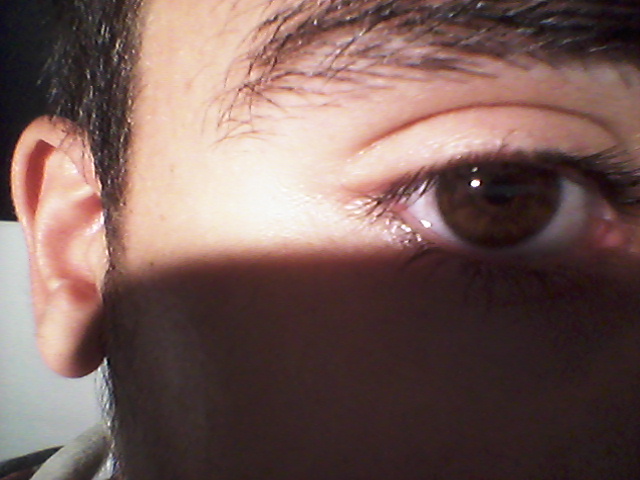} 
}\hspace{0.3cm}
\subfloat[]
{ 
    \includegraphics[scale=.03]{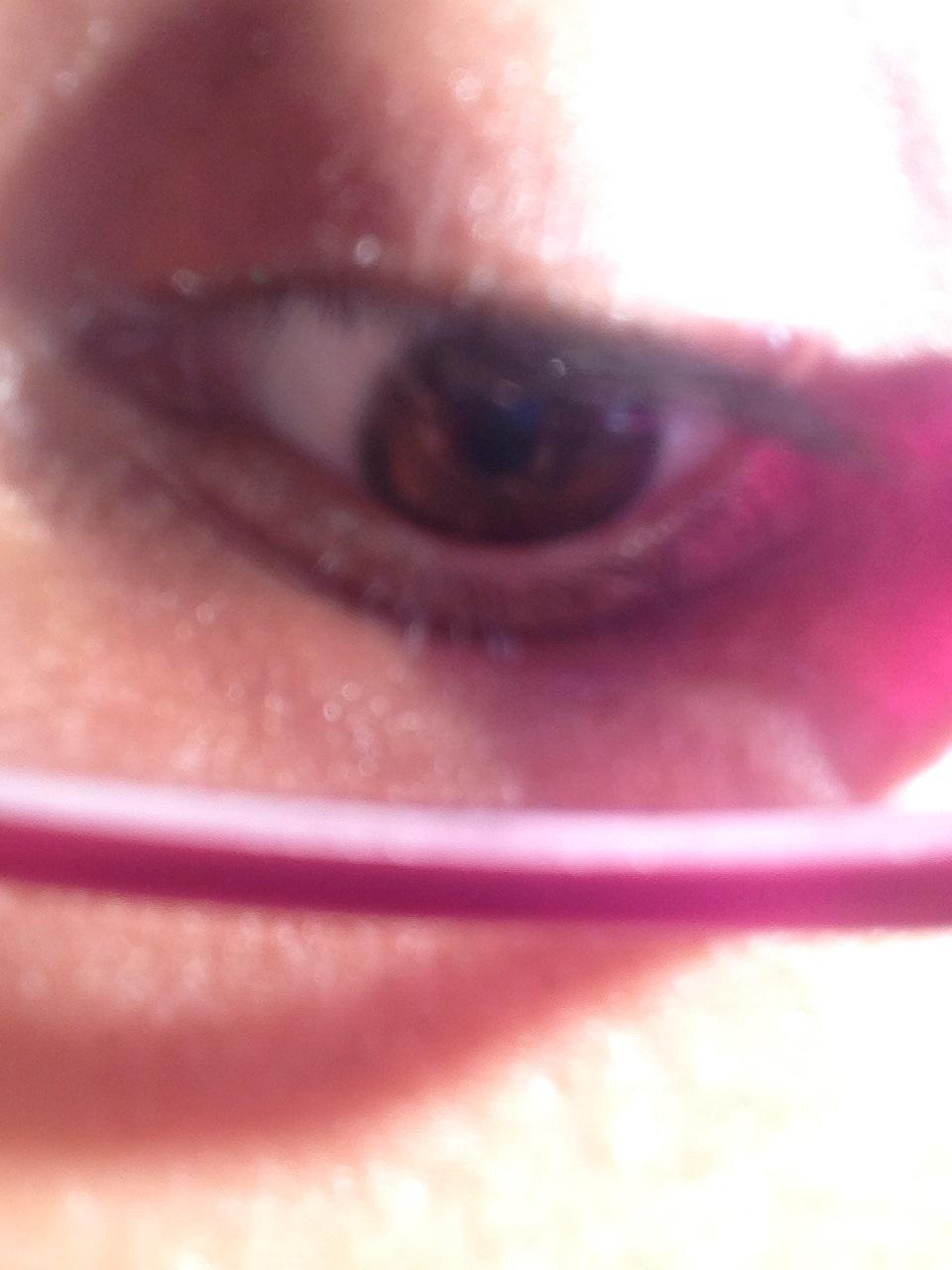} 
}\hspace{0.3cm}
\subfloat[]
{ 
    \includegraphics[scale=.03]{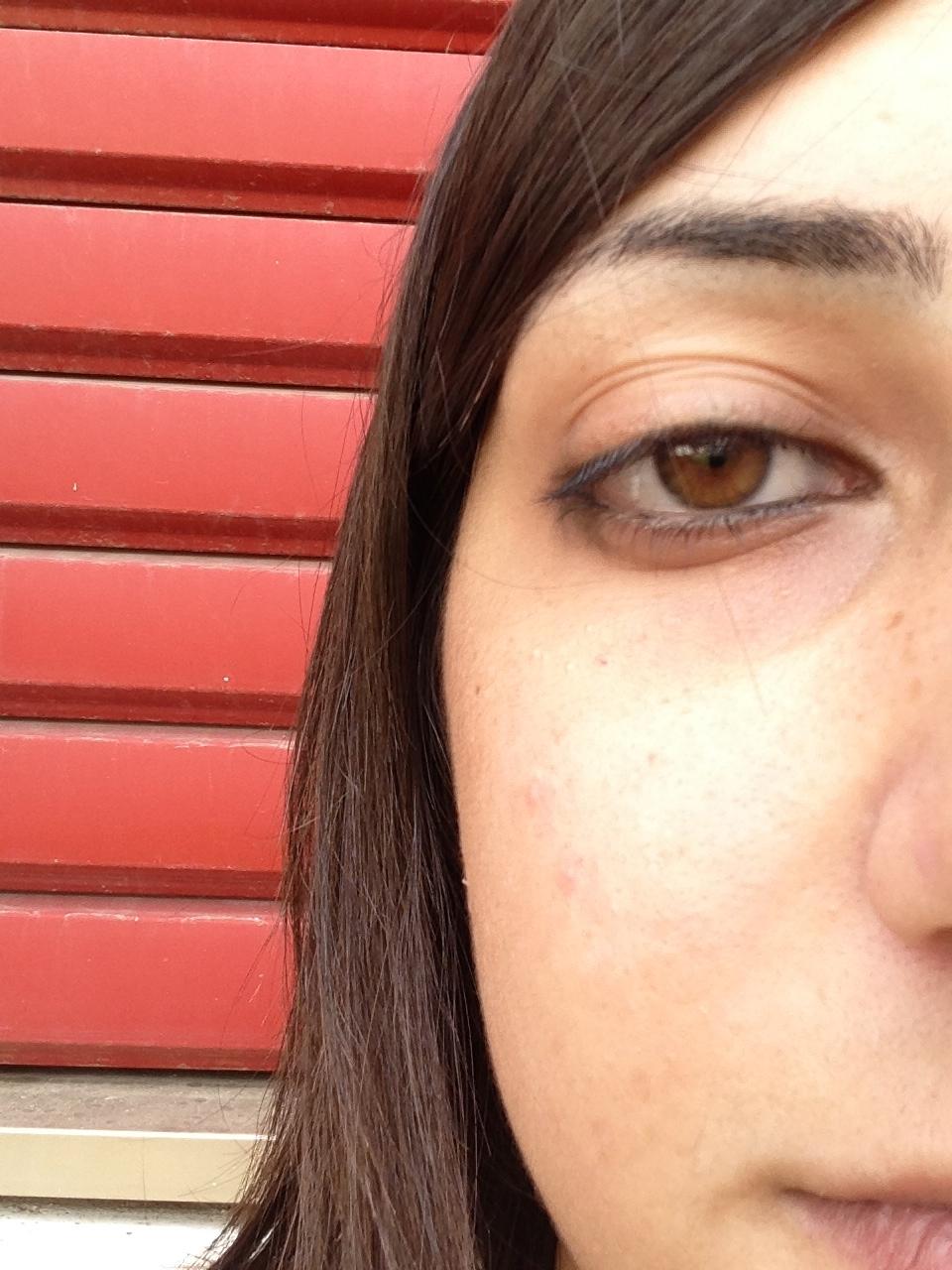}   
}\hspace{0.3cm}
\subfloat[]
{ 
    \includegraphics[scale=.07]{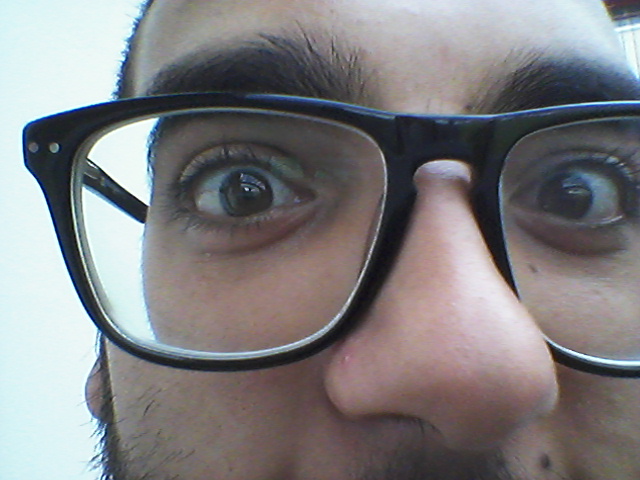}   
}

\caption{Example images from the challenging MICHE-I dataset. (a) Occlusion, (b) Downward gaze and specular reflection, (c) Prominent background in the outdoor setting and (d) A single image containing both eyes but labeled as right eye image (061\textunderscore GT2\textunderscore OU\textunderscore F\textunderscore RI\textunderscore 01\textunderscore 3 where, RI indicates right eye).}
\label{Fig: MICHE}
\end{figure}

In both cases of joint identification and joint verification experiments, we observed that the performance of the proposed method evaluated on the MICHE-I dataset was relatively worse compared to the remaining two datasets. We hypothesize that the poor performance can be attributed to two reasons: (i) the image characteristics, and (ii) the variation in the performance across different lateralities, \textit{i.e.,} left vs. right periocular images. MICHE-I dataset was assembled as a part of an iris challenge evaluation and contains images acquired in unconstrained settings (indoor and outdoor settings) having occlusions (specular reflection and downward gaze). See some challenging images from the MICHE-I dataset images in Figure~\ref{Fig: MICHE}. In contrast, CASIA and OULU datasets contain images acquired in controlled settings.

\begin{figure}[h]
\centering
\subfloat[]
{   
    \includegraphics[scale=.3]{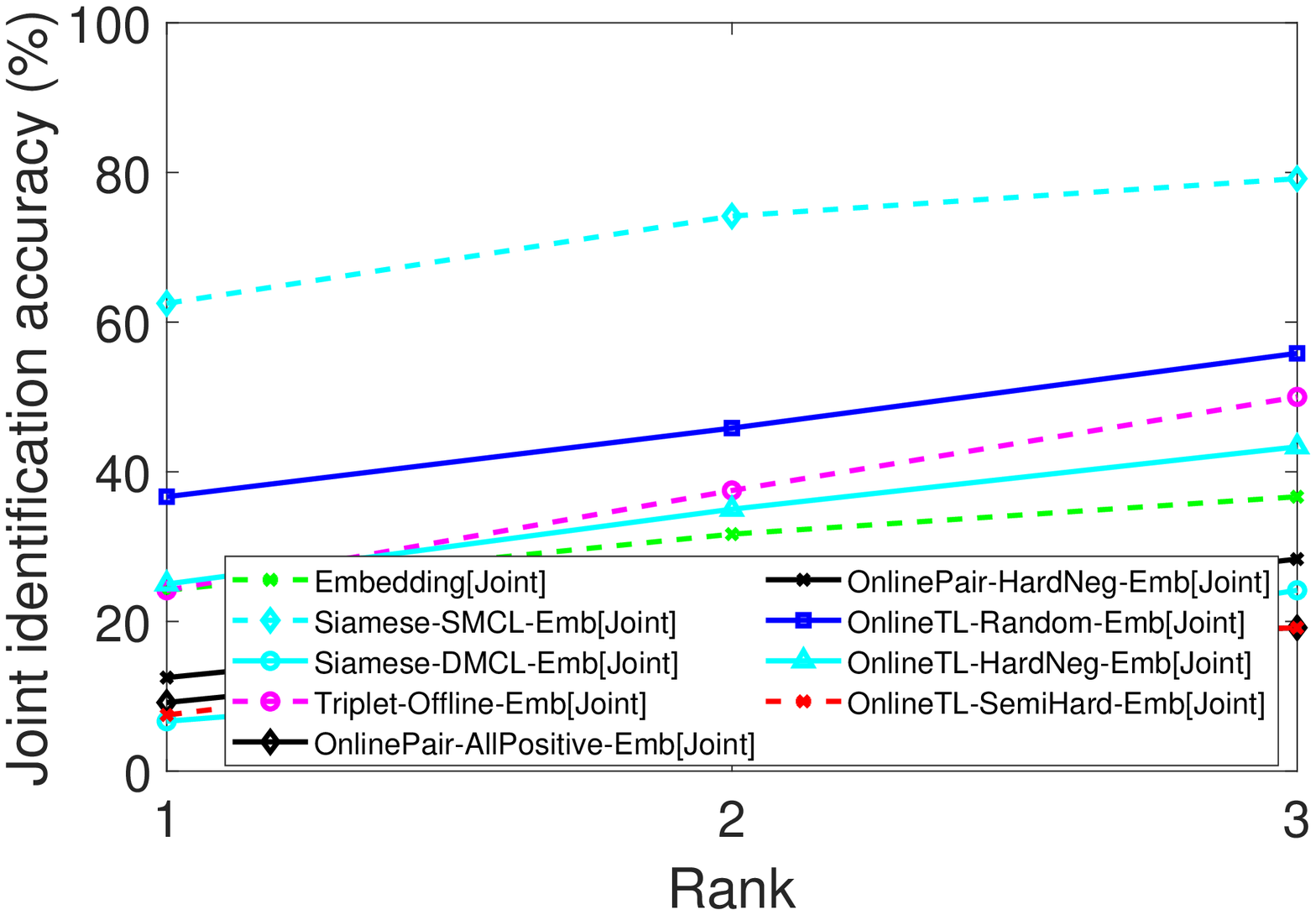} 
} \hspace{-0.75cm}
\subfloat[]
{ 
    \includegraphics[scale=.3]{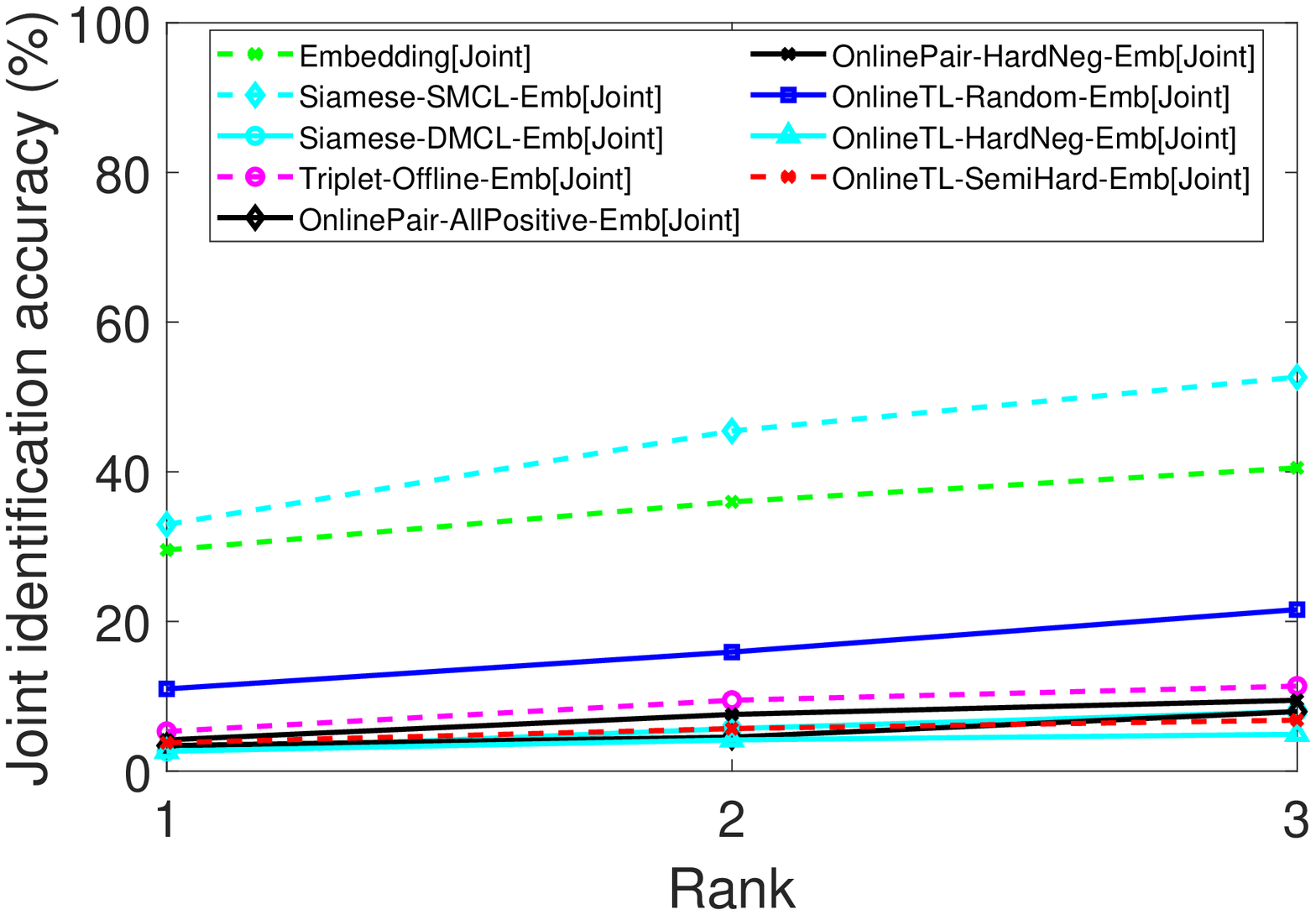} 
} 

\caption{Cumulative Matching Characteristics (CMC) curves for the proposed method in the joint identification scenario for the MICHE-I dataset evaluated separately on the two lateralities. (a) Left periocular images and (b) Right periocular images. Results indicate that the proposed method performs better on left periocular images compared to right periocular images.}
\label{Fig: CMCMICHE}
\end{figure}

We presented the CMC curves corresponding to joint identification results for two lateralities separately in Figure~\ref{Fig: CMCMICHE}. Results indicate that the proposed method performed better on left periocular images compared to right periocular images. This variation in the performance across the two lateralities resulted in the overall poor performance on the entire MICHE-I dataset. MICHE-I dataset has an imbalanced distribution of lateralities. Only 30\% of the total number of images are of left periocular images. We hypothesize that the imbalanced distribution coupled with some mislabeled test case (see Figure~\ref{Fig: MICHE}(d)) may have further compounded the challenges, resulting in overall poor performance.     

The \textbf{main findings} from the experiments are as follows: 
\begin{enumerate}
\item The joint biometric and sensor representation performed well in both joint identification scenario, with an average identification accuracy of $\sim80\%$ computed at Rank 1, and an average joint verification accuracy of $96\%$ at a false match rate of 5\%, averaged across the three biometric modalities. 
\item The representation is \textit{robust} across three modalities (iris, face and periocular), and different sensors (near-infrared iris sensors and visible smartphone sensors). 
\item The joint embedding outperformed baselines that used state-of-the-art commercial biometric matchers and sensor identification schemes across three datasets corresponding to three biometric modalities and multi-spectral sensors.  
\end{enumerate}

\section{Summary and Future Work}
\label{Sec:Sum}
In this paper, we proposed a one-shot method to simultaneously authenticate the user and the device from a single image, say a face or an iris image. To accomplish this task, we developed a method to learn a joint representation that can be used for combined biometric and sensor (device) recognition. The joint representation will be useful in remote application scenarios that employ multiple factor authentication. Examples include mobile banking on smartphones and accessing secure services. Additionally, the joint representation will be implicitly privacy-preserving as the biometric and sensor representations cannot be trivially separated. We used an embedding network to generate the joint representation. We evaluated the proposed approach by (i) exploring different training schemes, and (ii) testing the method on multiple datasets belonging to three different biometric modalities (iris, face and periocular) in both the identification and verification scenarios. We observed best performing results of identification accuracy of 99.81\% at Rank1 and a verification accuracy of TMR=100\% at 1\% FMR using the proposed method.

Future work will involve improving the current framework by incorporating domain knowledge (sensor characteristics) to further increase the recognition performances. We will also evaluate the proposed method on datasets with larger number of subjects and sensors.  

\balance
\bibliographystyle{ieeetran}
\bibliography{ICPR}

\begin{thebibliography}{10}
\providecommand{\url}[1]{#1}
\csname url@samestyle\endcsname
\providecommand{\newblock}{\relax}
\providecommand{\bibinfo}[2]{#2}
\providecommand{\BIBentrySTDinterwordspacing}{\spaceskip=0pt\relax}
\providecommand{\BIBentryALTinterwordstretchfactor}{4}
\providecommand{\BIBentryALTinterwordspacing}{\spaceskip=\fontdimen2\font plus
\BIBentryALTinterwordstretchfactor\fontdimen3\font minus
  \fontdimen4\font\relax}
\providecommand{\BIBforeignlanguage}[2]{{%
\expandafter\ifx\csname l@#1\endcsname\relax
\typeout{** WARNING: IEEEtran.bst: No hyphenation pattern has been}%
\typeout{** loaded for the language `#1'. Using the pattern for}%
\typeout{** the default language instead.}%
\else
\language=\csname l@#1\endcsname
\fi
#2}}
\providecommand{\BIBdecl}{\relax}
\BIBdecl

\bibitem{BiomSensor_1}
N.~{Bartlow}, N.~{Kalka}, B.~{Cukic}, and A.~{Ross}, ``Identifying sensors from
  fingerprint images,'' in \emph{IEEE Computer Society Conference on Computer
  Vision and Pattern Recognition Workshops}, 2009, pp. 78--84.

\bibitem{BiomSensor_2}
S.~Prabhakar, A.~Ivanisov, and A.~Jain, ``Biometric recognition: Sensor
  characteristics and image quality,'' \emph{Instrumentation \& Measurement
  Magazine, IEEE}, vol.~14, pp. 10 -- 16, 07 2011.

\bibitem{Galdi_PRL_15}
C.~Galdi, M.~Nappi, and J.~L. Dugelay, ``Multimodal authentication on
  smartphones: {C}ombining iris and sensor recognition for a double check of
  user identity,'' \emph{Pattern Recognition Letters}, vol.~3, pp. 34--40,
  2015.

\bibitem{Ross_ICB_18}
R.~Arjona, M.~A. Prada-Delgado, I.~Baturone, and A.~Ross, ``Securing minutia
  cylinder codes for fingerprints through physically unclonable functions: An
  exploratory study,'' in \emph{Proc. of 11th {IAPR} International Conference
  on Biometrics ({ICB})}, Gold Coast, Australia, June 2018.

\bibitem{Jain_Intro}
A.~K. Jain, A.~Ross, and K.~Nandakumar, ``Introduction to biometrics,''
  \emph{Springer}, 2011.

\bibitem{Lukas_TIFS_06}
J.~Lukas, J.~Fridrich, and M.~Goljan, ``Digital camera identification from
  sensor pattern noise,'' \emph{IEEE Transactions on Information Forensics and
  Security}, vol.~1, no.~2, pp. 205--214, June 2006.

\bibitem{Galdi_ICIAP}
C.~Galdi, M.~Nappi, and J.-L. Dugelay, ``Combining hardwaremetry and biometry
  for human authentication via smartphones,'' in \emph{Image Analysis and
  Processing ({ICIAP})}, V.~Murino and E.~Puppo, Eds.\hskip 1em plus 0.5em
  minus 0.4em\relax Springer International Publishing, 2015, pp. 406--416.

\bibitem{Galdi_Face}
------, ``Secure user authentication on smartphones via sensor and face
  recognition on short video clips,'' in \emph{Green, Pervasive, and Cloud
  Computing}, M.~H.~A. Au, A.~Castiglione, K.-K.~R. Choo, F.~Palmieri, and
  K.-C. Li, Eds.\hskip 1em plus 0.5em minus 0.4em\relax Springer International
  Publishing, 2017, pp. 15--22.

\bibitem{Chen_TIFS_08}
M.~Chen, J.~Fridrich, M.~Goljan, and J.~Lukas, ``Determining image origin and
  integrity using sensor noise,'' \emph{IEEE Transactions on Information
  Forensics and Security}, vol.~3, no.~1, pp. 74--90, March 2008.

\bibitem{BTAS_19}
S.~Banerjee and A.~Ross, ``Smartphone camera de-identification while preserving
  biometric utility,'' in \emph{Proc. of 10th {IEEE} International Conference
  on Biometrics: Theory, Applications and Systems ({BTAS})}, Tampa, USA,
  September 2019.

\bibitem{Uhl4_ICB_12}
A.~Uhl and Y.~H{\"o}ller, ``Iris-sensor authentication using camera {PRNU}
  fingerprints,'' in \emph{5th IAPR International Conference on Biometrics
  ({ICB})}, March 2012, pp. 230--237.

\bibitem{Emb1}
L.~Wang, Y.~Li, and S.~Lazebnik, ``Learning deep structure-preserving
  image-text embeddings,'' \emph{IEEE Conference on Computer Vision and Pattern
  Recognition (CVPR)}, pp. 5005--5013, 2016.

\bibitem{CasV2}
``C{ASIA} {I}ris {D}atabase {V}ersion 2,'' \url{
  http://biometrics.idealtest.org/dbDetailForUser.do?id=2}, [Online accessed:
  18th December 2019].

\bibitem{MICHE}
M.~D. Marsico, M.~Nappi, F.~Narducci, and H.~Proen{\c c}a, ``Insights into the
  results of {MICHE I} - mobile iris challenge evaluation,'' \emph{Pattern
  Recognition}, vol.~74, pp. 286 -- 304, 2018.

\bibitem{OULU}
Z.~Boulkenafet, J.~Komulainen, L.~Li, X.~Feng, and A.~Hadid, ``{OULU-NPU}: A
  mobile face presentation attack database with real-world variations,''
  \emph{IEEE International Conference on Automatic Face and Gesture
  Recognition}, 2017.

\bibitem{siamese}
J.~Bromley, I.~Guyon, Y.~LeCun, E.~S\"{a}ckinger, and R.~Shah, ``Signature
  verification using a "siamese" time delay neural network,'' in \emph{Proc. of
  6th International Conference on Neural Information Processing Systems}, 1993,
  pp. 737--744.

\bibitem{contrastive}
S.~Chopra, R.~Hadsell, and Y.~Lecun, ``Learning a similarity metric
  discriminatively, with application to face verification,'' \emph{IEEE
  Conference on Computer Vision and Pattern Recognition (CVPR)}, vol.~1, pp.
  539-- 546, 07 2005.

\bibitem{tripletnetwork}
E.~Hoffar and N.~Ailon, ``Deep metric learning using triplet network,'' in
  \emph{International Workshop on Similarity-Based Pattern Recognition}.\hskip
  1em plus 0.5em minus 0.4em\relax Springer, 2015, pp. 84--92.

\bibitem{tripletloss}
M.~Schultz and T.~Joachims, ``Learning a distance metric from relative
  comparisons,'' in \emph{Proceedings of the 16th International Conference on
  Neural Information Processing Systems}, S.~Thrun, L.~K. Saul, and
  B.~Sch\"{o}lkopf, Eds., 2003, pp. 41--48.

\bibitem{TLmining}
F.~{Schroff}, D.~{Kalenichenko}, and J.~{Philbin}, ``{FaceNet}: A unified
  embedding for face recognition and clustering,'' in \emph{IEEE Conference on
  Computer Vision and Pattern Recognition (CVPR)}, June 2015, pp. 815--823.

\bibitem{npair}
K.~Sohn, ``{Improved Deep Metric Learning with Multi-class N-pair Loss
  Objective},'' in \emph{Proceedings of the 30th International Conference on
  Neural Information Processing Systems}, D.~D. Lee, M.~Sugiyama, U.~V.
  Luxburg, I.~Guyon, and R.~Garnett, Eds., 2016, pp. 1857--1865.

\bibitem{imp}
``Embedding network tutorial,''
  \url{https://github.com/adambielski/siamese-triplet/blob/master/Experiments_MNIST.ipynb},
  [Online accessed: 12th January, 2020].

\bibitem{CR_Ref1}
F.~Marra, G.~Poggi, C.~Sansone, and L.~Verdoliva, ``A deep learning approach
  for iris sensor model identification,'' \emph{Pattern Recognition Letters},
  vol. 113, pp. 46 -- 53, 2018.

\bibitem{CR_Ref2}
D.~Freire-Obregón, F.~Narducci, S.~Barra, and M.~Castrillón-Santana, ``Deep
  learning for source camera identification on mobile devices,'' \emph{Pattern
  Recognition Letters}, vol. 126, pp. 86 -- 91, 2019.

\bibitem{CR_Ref3}
A.~Agarwal, R.~Keshari, M.~Wadhwa, M.~Vijh, C.~Parmar, R.~Singh, and M.~Vatsa,
  ``Iris sensor identification in multi-camera environment,'' \emph{Information
  Fusion}, vol.~45, pp. 333 -- 345, 2019.

\bibitem{Li_TIFS_10}
C.~T. Li, ``Source camera identification using enhanced sensor pattern noise,''
  \emph{IEEE Transactions on Information Forensics and Security}, vol.~5,
  no.~2, pp. 280--287, June 2010.

\bibitem{Ross_18}
S.~{Banerjee} and A.~{Ross}, ``Impact of photometric transformations on {PRNU}
  estimation schemes: {A} case study using near infrared ocular images,'' in
  \emph{International Workshop on Biometrics and Forensics (IWBF)}, June 2018,
  pp. 1--8.

\bibitem{PRNU_JPEG}
D.~{Valsesia}, G.~{Coluccia}, T.~{Bianchi}, and E.~{Magli}, ``User
  authentication via {PRNU}-based physical unclonable functions,'' \emph{IEEE
  Transactions on Information Forensics and Security}, vol.~12, no.~8, pp.
  1941--1956, 2017.

\bibitem{ResNet}
K.~Hernandez-Diaz, F.~Alonso-Fernandez, and J.~Bigun, ``Periocular recognition
  using {CNN} features off-the-shelf,'' in \emph{Proc. of 17th International
  Conference of the Biometrics Special Interest Group (BIOSIG)}, September
  2018, pp. 1--5.

\bibitem{TSNE}
L.~van~der Maaten and G.~E. Hinton, ``Visualizing data using t-{SNE},''
  \emph{Journal of Machine Learning Research}, vol.~9, pp. 2431--2556, November
  2008.

\end{thebibliography}

\end{document}